\def\year{2020}\relax
\documentclass[letterpaper]{article} 
\usepackage{aaai21}  
\usepackage{times}  
\usepackage{helvet} 
\usepackage{courier}  
\usepackage[hyphens]{url}  
\usepackage{graphicx} 
\usepackage{booktabs}
\usepackage[switch]{lineno}
\usepackage{natbib}  
\usepackage{caption} 
\title{Delaying Interaction Layers in Transformer-based Encoders for Efficient Open Domain Question Answering}
\author{

    Wissam Siblini, Mohamed Challal, Charlotte Pasqual 
}
\affiliations{

    Worldline, France
}

\begin{document}
\maketitle
\begin{abstract}
Open Domain Question Answering (ODQA) on a large-scale corpus of documents (e.g. Wikipedia) is a key challenge in computer science. Although transformer-based language models such as Bert have shown on SQuAD the ability to surpass humans for extracting answers in small passages of text, they suffer from their high complexity when faced to a much larger search space. The most common way to tackle this problem is to add a preliminary Information Retrieval step to heavily filter the corpus and only keep the relevant passages. In this paper, we propose a more direct and complementary solution which consists in applying a generic change in the architecture of transformer-based models to delay the attention between subparts of the input and allow a more efficient management of computations. The resulting variants are competitive with the original models on the extractive task and allow, on the ODQA setting, a significant speedup and even a performance improvement in many cases.
\end{abstract}

The last few years have given rise to many disruptive innovations in the field of Natural Language Processing (NLP) and allowed a significant improvement on public benchmarks \cite{wolf2019huggingfaces}. In particular, the proposal of a novel architecture called the \textit{Transformer} \cite{vaswani2017attention} and an adaptation into the versatile easy-to-use language model Bert \cite{devlin2019bert} have led to a series of publications generating a continual enthusiasm. Recent transformer-based models such as RoBerta \cite{liu2019roberta}, XLNet \cite{yang2019xlnet}, Albert \cite{lan2019albert}, Electra \cite{clark2019electra}, T5 \cite{raffel2019exploring} managed to outperform humans on difficult tasks on GLUE \cite{wang2018glue}, the benchmark for general language comprehension assessment. These exploits quickly led to the democratization of their use in many applications. We here focus on automatic question answering where we search for the answer of a user question in a large set of text documents (e.g. the entire English Wikipedia with millions of articles). Language models have been proven efficient on a sub-task called extractive Question Answering (eQA), sometimes also referred to as Reading Comprehension (RC), on the reference SQuAD dataset \cite{rajpurkar2016squad}: it is made of question-document pairs and the goal is to find the answer within the document. But on our target task, Open Retrieval/Domain Question Answering (ODQA), the problem is more complex because for each question the search space is much larger. This entails a complexity issue: transformer-based readers require a non negligible time to process a single question-paragraph pair so they are not able to manage millions in real-time. The most common solution is to combine eQA with Information Retrieval (IR) \cite{manning2008introduction} to first select $p$ relevant documents and only apply the costly reading comprehension model on them. Such a combination has proven itself in BertSerini \cite{yang2019end} where the widely known Lucene with BM25 \cite{robertson1995okapi,bialecki2012apache} for the IR part was combined with Bert for the eQA part. 
In this paper, we propose to tackle the time issue from a more direct and complementary angle which consists in adapting the structure of the eQA model so that many computations can be saved or only done once as a preprocessing. More precisely, our contributions are the following:
\begin{itemize}
\item We use a Delaying Interaction Layers mechanism (DIL) on transformer-based models that consists in only applying the attention between subparts (segments) of the input sequence in the last blocs of the architecture. This allows an efficient management of the computations for Open Domain Question Answering. We implement this mechanism for both Bert and Albert and refer to our variants as DilBert and DilAlbert.
\item We study the behavior of the variants in the standard eQA setting and show that they are both competitive with the base models.
\item We also analyze the impact of delayed interaction on the models complexity and then empirically confirm that in the ODQA setting, it allows to speed up computations by up to an order of magnitude on either GPU or CPU.
\item  Finally, we evaluate the models on the reference ODQA dataset OpenSQuAD \cite{chen2017reading} by combining them with Answerini as \citet{yang2019end}. Although DilBert (resp. DilAlbert) performs slightly worse than Bert (resp. Albert) when faced to a single relevant passage, it can outperform it in the ODQA setting when having to select the right answer within several paragraphs. 
\end{itemize}

Moreover, we report updated and improved results on OpenSQuAD for the simple baseline Lucene + Bert Base English uncased, due to engineering improvement.

\section{Related Works}

For a long time, researchers have been interested in Automatic Question Answering \cite{woods1977lunar} to build intelligent search engines and browse large-scale unstructured documents databases such as Wikipedia \cite{chen2017reading,ryu2014open,wang2017r,yang2019end,lee2018ranking}. This kind of system is often designed with several layers that successively select a smaller but more precise piece of text. It generally starts with a one-stage or a multiple-stage \textit{Information Retrieval} step (Ad-hoc retrieval) that identifies relevant documents for the question at hand. Then, an algorithm designed for \textit{eQA} is applied to identify the answer spans within the selected passages.

\subsection{Ad-hoc Retrieval}

Consider a user query that reflects a need for information. To look for relevant items within a large set of documents, web pages, the general approach consists in: (i) applying an encoding model to all documents, (ii) encoding the query as well, (iii) applying a ranker that produces a relevance score for each query-document pair based on their encodings, and finally (iv) sorting the documents accordingly \cite{manning2008introduction,mitra2018introduction,buttcher2016information}. For the encoding part, proposals go from vocabulary and frequency based strategies such as bag-of-words or TF-IDF \cite{baeza1999modern,hiemstra2000probabilistic}, to word2vec embeddings averaging \cite{mikolov2013distributed,ganguly2015word} to even strongly contextualized embeddings with Bert \cite{devlin2019bert,macavaney2019cedr,nogueira2019passage,yang2019simple}. For the relevance score part, the most popular choice is a fixed similarity/distance measure (cosine or Euclidean) but researchers have explored others strategies such as siamese networks \cite{das2016together}, histograms \cite{guo2016deep}, convolutional networks \cite{dai2018convolutional}, etc. In the recent litterature \cite{macavaney2019cedr}, several combinations between contextualized embedding techniques (Elmo and Bert) and rankers (DRMM, KNRM, PACRR) have been tried. By fine-tuning BERT for IR and exploiting the representation of its [CLS] token, the authors obtained state-of-the-art results on several reference IR benchmarks. 

On another side, there is an old and proven IR baseline called BM25 \cite{robertson1995okapi}, a weighted similarity function based on TF-IDF statistics, which until today remains very competitive with neural methods \cite{lin2019neural}, and moreover is extremely fast and benefits from decades of fine engineering, for instance in its implementation in Lucene \cite{yang2018anserini,yang2017anserini,bialecki2012apache}.

\subsection{Extractive Question Answering}

\begin{figure}[t]
\centering
\includegraphics[width=0.87\columnwidth]{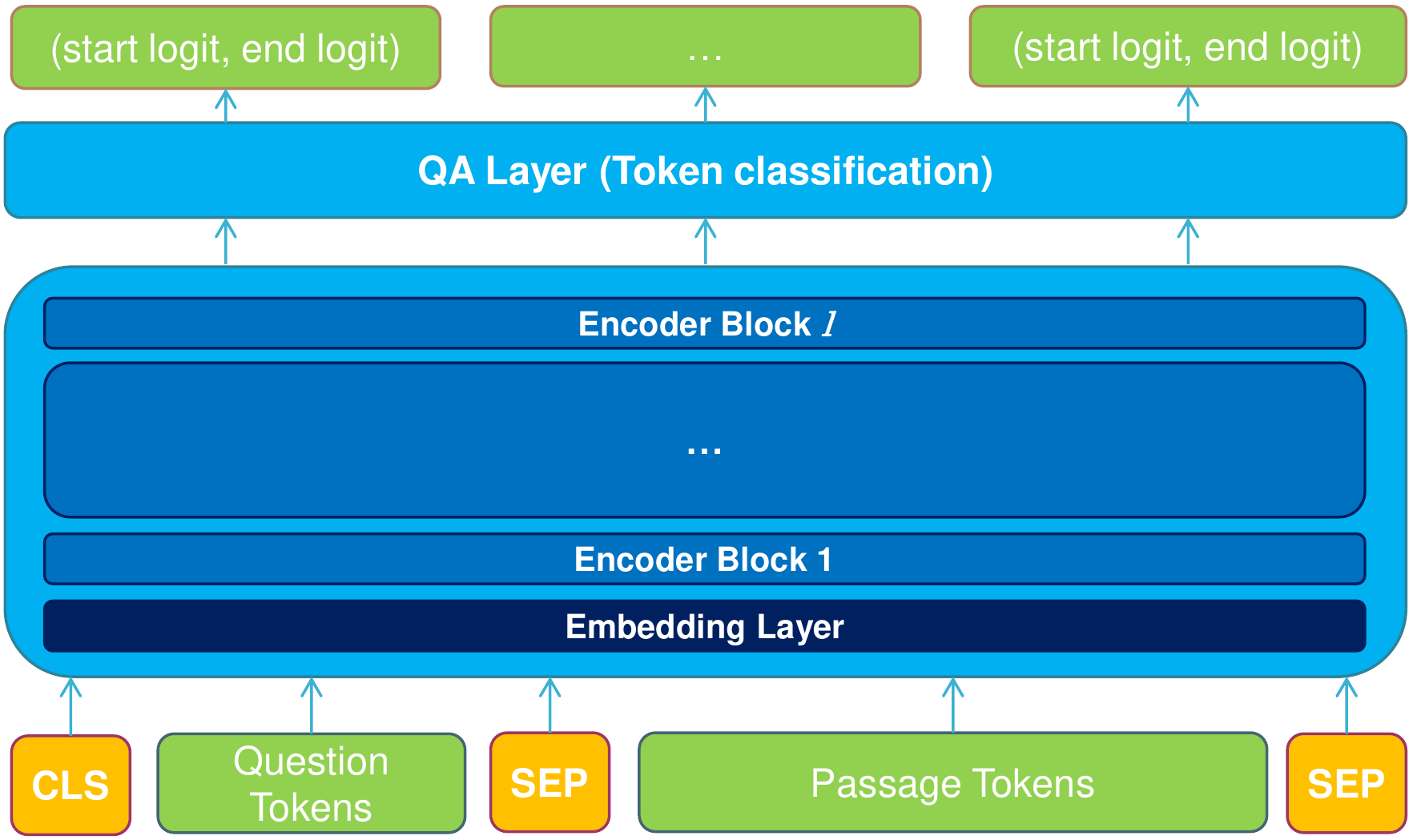} 
\caption{General architecture of transformer-based language models ($l$ blocs) designed for extractive Question Answering. Start logit (resp. end logit) refers to a relevance score predicted by the model for each token to be the beginning (resp. the end) of the answer.}
\label{encoder_for_qa}
\end{figure}

The extractive Question Answering task consists in identifying a question's answer as a text span within a rather small passage. The most popular dataset for this task is the Stanford Question Answering Dataset (SQuAD) \cite{rajpurkar2016squad} which consists in more than a hundred thousand questions, each paired with a Wikipedia article paragraph. The state of art algorithms for this task are transformer-based language models such as Bert \cite{devlin2019bert} or Albert \cite{lan2019albert}. They are usually composed of an input embedding layer, followed by a succession of $l$ encoder blocs (which implement, in particular, a self-attention mechanism) and finally an output layer. On the eQA task, they take as input a text sequence that is the concatenation of the question and the associated passage. The whole sequence is tokenized and a special [CLS] (resp. [SEP]) token is added at the beginning (resp. between the question and the passage and at the end). Then, the associated features (token ids, segments, positions...) are fed to the model which predicts two probabilities for each token to be the beginning and the end of the answer span (Figure \ref{encoder_for_qa}).  

\subsection{Open Domain Question Answering}

Open Domain Question answering solutions generally combine a retriever that solves the Ad-hoc Retrieval task by selecting relevant documents and a reader that solves the extractive Question Answering task on the selected passages. An emblematic example is DrQA \cite{chen2017reading}, a bot developed at Facebook which is able to answer to questions by searching in the entire English Wikipedia in real-time. It consists in a TF-IDF + cosine retriever that selects 5 relevant documents followed by a multi-layer RNN reader \cite{chen2016thorough}. Later, other proposals were able to surpass its quality of answer with a paragraph reranker \cite{lee2018ranking,wang2017r} and then with a "minimal" retriever that selects small but relevant portions of text \cite{min2018efficient}. In the meantime, Bert was released and \citet{yang2019end} proposed a first successful usage for ODQA, by simply combining answerini/Lucene with a Bert base fine-tuned on SQuAD, which achieved the new state-of-the art performance on the reference benchmark OpenSQuAD \cite{chen2017reading}. Since then, a few proposals \cite{lee2019latent,feldman2019multi,wang2019multi, ren2020multi} have been focusing on better selection of passages and a better pooling of answers from them.

In all of these proposals, the scaling issue is always tackled from the retriever's perspective. Yet, the ODQA setting offers many opportunities to address it from the reader's side, which is the main motivation behind this work. 

\section{Delaying Interaction Layers}
\label{sec:proposal}
In the ODQA setting, the set of documents is rather static. If $q$ questions are asked and there are $p$ documents in the database, without any retrieval step, transformer-based readers will consider all the $q\times p$ possible pairs, split them into fixed-length examples and run inference on them. Not only this is very costly but all computations are done interactively. More precisely, in the encoder layers, the attention mechanism makes the representation of the documents tokens dependent on the question tokens, so there is no possibility to pre-compute intermediate representations of the database's documents without knowing the question beforehand.

\paragraph{Key Idea}

To better face this setting, we consider a structural modification of existing models to allow reducing computations and make preprocessing possible. The key idea is to "delay" the interaction between question and paragraph so that a part of the computations can be done independently. Also, we limit ourselves to (i) changes that are not specific to a given model so that the proposal can be applied to a maximum of transformer-based language models and (ii) do not introduce new parameters so that we can benefit from existing pre-trained weights. 
 
\paragraph{Algorithm and Variants}

\begin{figure}[t]
\centering
\includegraphics[width=0.97\columnwidth]{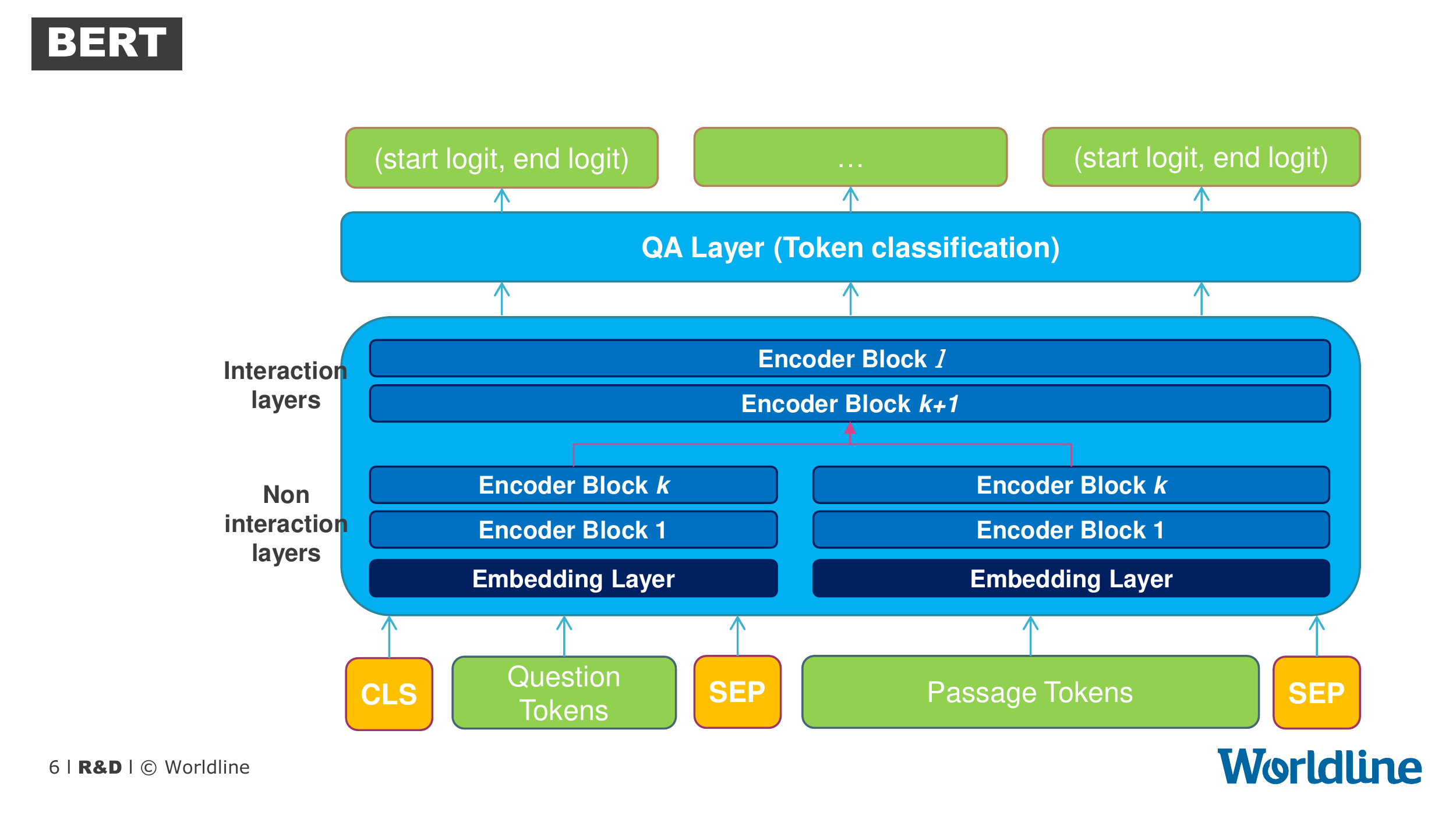} 
\caption{Structural change in the transformer-based architecture (see original one in Figure \ref{encoder_for_qa}): here the first $k$ non interaction layers/blocs are applied independently to question and paragraph and the last $l-k$ interaction layers are applied to the whole.}
\label{dil_encoder_for_qa}
\end{figure}

We start by splitting the input between its two segments after the separator [SEP]. We apply the input layer and the first $k$ encoder blocs independently on both parts, then concatenate the two outputs and finally apply the last $l - k$ encoder blocs and the output layer (Figure \ref{dil_encoder_for_qa}). We refer to the first $k$ blocs as the non interaction layers and the last $l-k$ as the interaction layers. The impact of hyperparameter $k$ is studied in the experimental section.

Several variants can be considered when fine-tuning a model with the delayed interaction framework: non interaction layers, applied to the question or to the paragraph, can either have the same weights or not; for specific cases such as Albert in which the $l$ blocs share their weights, the weights of the first $k$ blocs and the last $l-k$ blocs can either evolve independently or not. In the following, we will consider the variant that does not increase the number of weights as compared to the original model. Our proposal is implemented in Python and is based on original models' implementation from the \textit{transformers} python library, version 2.7.0 \cite{wolf2019huggingfaces}.

\paragraph{Complexity}

Delayed interaction reduces the complexity of the first $k$ blocs. Specifically, the self-attention mechanism requires a number of computations proportional to $n_s^2$ where $n_s$ is the input sequence length. $n_s$ can be decomposed into the sum of the question length $n_q$ (including the [CLS] and the first [SEP]) and the paragraph length $n_p$ (including the last [SEP]). Therefore, the complexity is reduced from $O(n_q^2 + n_p^2 + 2n_p\times n_q)$ to $O(n_q^2 + n_p^2)$.

Although it already makes a difference, this can be negligible especially when the paragraph length is much larger than the question length and when the hardware has a great parallelization power (e.g. a GPU). The big difference takes place in the ODQA setting. Let us consider that a single encoder bloc has a forward complexity of $C\times n_s^2$. Therefore, the whole encoder part in a transformer-based model has a complexity of $l\times C \times n_s^2$ to process a single example, and $l\times C \times q\times p\times n_s^2$ to process a set of $q$ questions and $p$ paragraphs (for the sake of simplicity, we assume here that paragraphs are cut such that question-paragraph pairs fit in the model input size $n_s$). With delayed interaction, the computations in the first $k$ blocs for each question (resp. paragraph) do not have to be redone for each paragraph (resp. question). Therefore, the complexity becomes $k \times C \times (q  \times n_q^2 + p \times n_p^2)  + (l-k) \times C \times q \times p \times n_s^2$. Even with $p$ and $q$ only of the order of $10^2$, the quadratic term with $p \times q$ largerly dominates. Also note that the paragraph processing in the first $k$ blocs (i.e. the term $k \times C \times p \times n_p^2$) can be done as an initialization step in the ODQA setting and does not need to be computed interactively at inference time when users ask questions. Therefore, delayed interaction entails a complexity decrease which roughly boils down to the ratio $\frac{l-k}{l}$ between the number of interaction layers in the Dil variant and the total number of layers in the original model.

\section{Experimental Study}

To assess the interest of delayed interaction, we run experiments on the reference Open Domain Question Answering challenge OpenSQuAD as \citet{yang2019end}, starting by the associated extractive Question Answering sub-task SQuAD v1.1 \cite{rajpurkar2016squad}. 
We consider two language models as baselines. The first one is Bert since it is the most used\footnote{\url{https://huggingface.co/models}} and also the one with the largest set of different pre-trained weights. And, to test our framework in a challenging way, the second one is Albert because it has the specificity of using the same weights in all of its $l$ encoder blocs and this could have an unexpected impact on the mechanism that we explore. We refer to our variant as DilBert (resp. DilAlbert) which stands for "Delaying Interaction Layers" in Bert (resp. Albert).

We start by evaluating DilBert/DilAlbert on the eQA sub-task to know how delayed interaction impacts the predictive performance with respect to the hyperparameter $k$. Then, we analyze the GPU/CPU speedup and the performance in the ODQA setting.

\subsection{Extractive QA}

\begin{figure}[t]
\centering
\includegraphics[width=0.87\columnwidth]{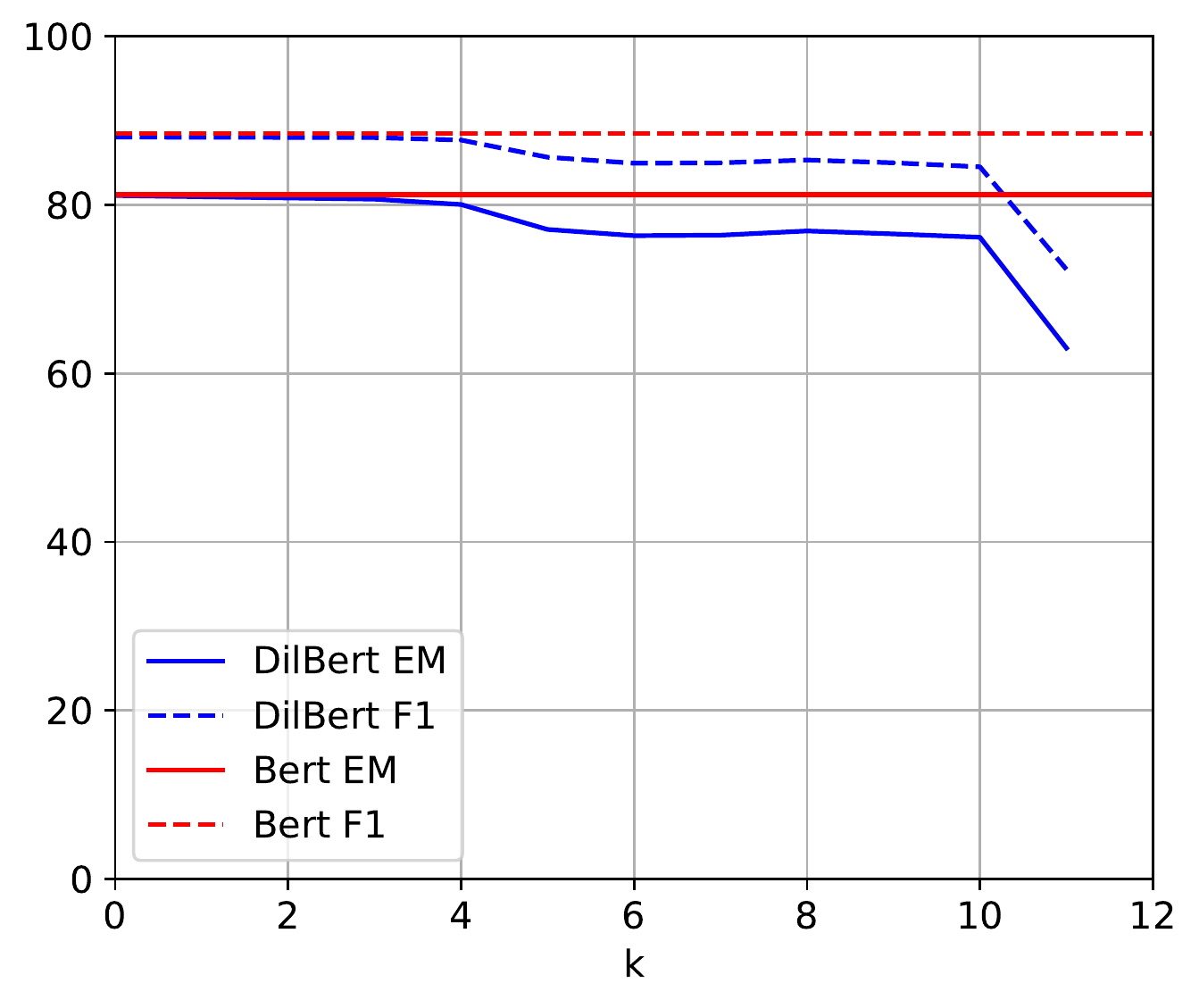} 
\caption{Evolution of DilBert's Exact Match (EM) and F1-score (F1) on the SQuAD v1.1 dev set with respect to the number of non interaction layers $k$. The performance of Bert is also displayed as a horizontal line.}.
\label{evo_performance_dilbert}
\end{figure}

\begin{figure}[t]
\centering
\includegraphics[width=0.87\columnwidth]{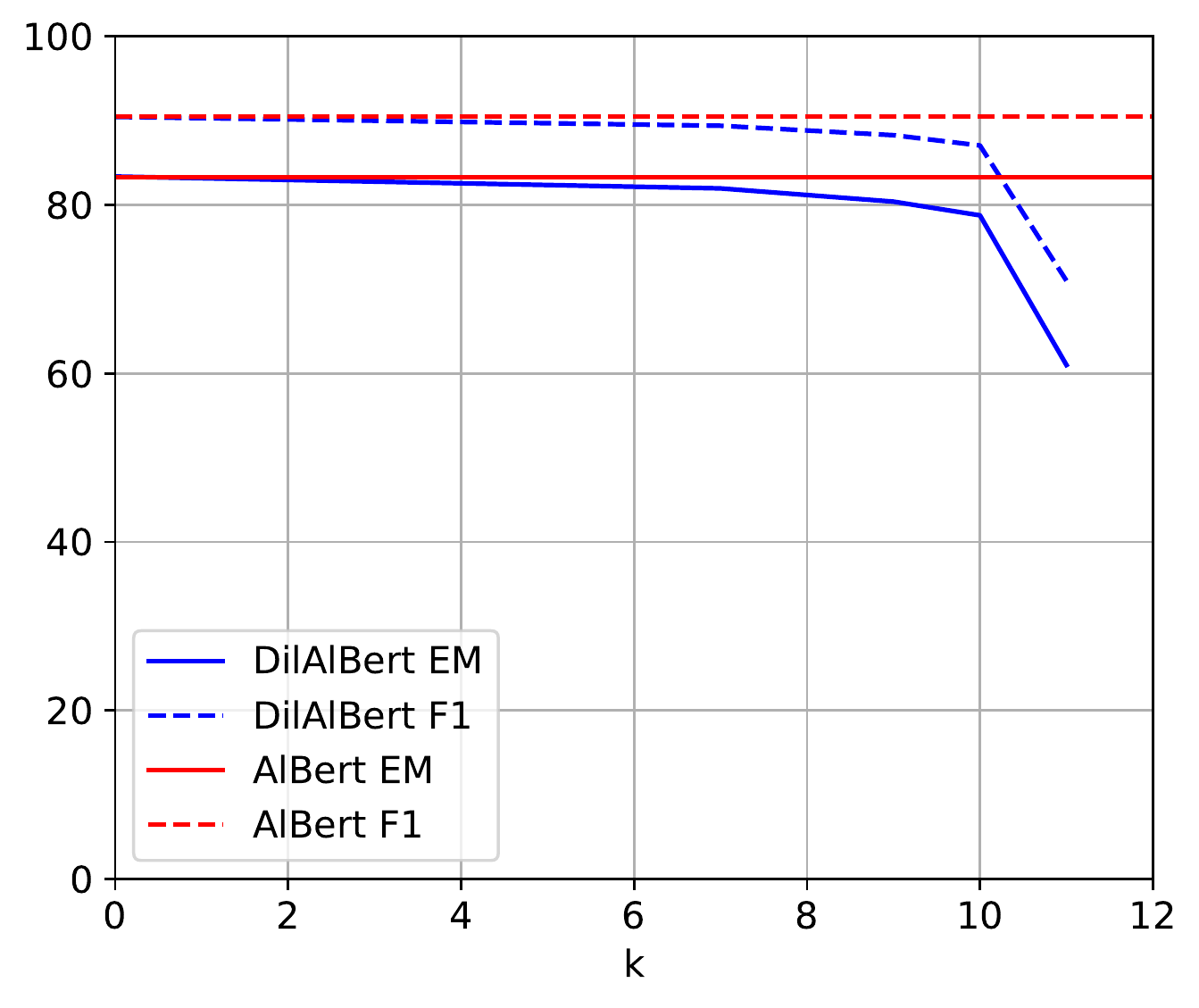}
\caption{Evolution of DilAlbert's Exact Match (EM) and F1-score (F1) on the SQuAD v1.1 dev set with respect to number of non interaction layers $k$. The performance of Albert is also displayed as a horizontal line.}
\label{evo_performance_dilalbert}
\end{figure}

For the eQA task, we train Bert, DilBert, Albert and DilAlbert on the SQuaD v1.1 train set. We consider the base version of each model and, as a starting point, we consider the English pre-trained weights: \textit{bert-based-uncased} for Bert and DilBert and \textit{albert-base-v2} for Albert and DilAlbert. 
The training (single run) is carried on with the \textit{run\_squad.py} script from the \textit{transformer} library using the default hyperparameters\footnote{\url{https://github.com/huggingface/transformers/tree/master/examples/question-answering}}: 2 epochs, an input sequence length $n_s$ of 384, a batch size of 12, and a learning rate of 3e-5. The models are then evaluated on the SQuAD v1.1 dev set using the two official metrics for the task (Exact Match and F1-score). For DilBert and DilAlbert, we make the hyperparameter $k$ vary from $0$ to $l-1$ and analyze how this impacts the performance (Figure \ref{evo_performance_dilbert} and \ref{evo_performance_dilalbert}).

As a sanity check, we can observe that without delayed interaction ($k=0$), our implementation of DilBert (resp. DilAlbert) provides the same results as Bert (resp. Albert). Then, as $k$ increases, the performance remains very competitive and it only decreases significantly for $k=11=l-1$. 
For $k=l$, there is no interaction anymore so start logits and end logits associated to paragraph tokens do not depend on the question and the performance dramatically drops to an exact match of around 13 (not reported in the Figure). We note a slight difference between the behavior of DilBert and DilAlbert. Whereas the latter seem to have a smooth evolution with respect to $k$, the former has two plateaus from $k=0$ to $k=4$ and from $k=5$ to $k=10$ and a decrease of performance in between. With inspiration from \citet{clark2019does}, our hypothesis is that since each bloc in Bert has its own set of weights, it could have been that the pre-trained attention heads in the $6^{th}$ layer had a component useful for the targeted eQA task. 

We also add that completely removing $k$ layers in the original models provide results significantly lower than the proposed Dil variants with $k$ non interaction layers. For instance, a 2-layer Bert, trained with the same protocol, obtains an exact match of 26.4 and an F1-score of 36.2 on the SQuAD v1.1 dev set.

Experiments are all carried on SQuAD v1.1 because OpenSQuAD is based on it. Nevertheless, the variants also apply, to some extent, to SQuAD v2.0. For instance, Albert, DilAlbert$_{k=6}$ and DilAlbert$_{k=10}$ respectively obtain an F1-Score of 81.4, 80.3 and 72.4 on its dev set. 

\subsection{Speedup}

To analyze the speedup entailed by delayed interaction, we consider a simple open domain question answering setting where $q = 100$ questions are asked and the RC model (e.g. Bert) searches for the answers in a database of $p = 100$ passages. We measure the total time required by Bert, Albert, DilBert and DilAlbert to process all question-paragraph pairs (Table \ref{time_consumption_encoder_dilencoder}). For the Dil variants, we detail the time required in the non interaction layers to do the independent processing on questions and paragraphs and in the interaction layers to do the dependent processing on question-paragraph pairs. We consider two values for the number of non interaction layers: $k=10$ because it allows preserving most of the eQA performance and $k=11$, the largest possible value here. 

\begin{table*}[t]
\centering
\resizebox{.95\textwidth}{!}{
\begin{tabular}{cccccccc}
\toprule
&  &  Bert  &  DilBert$_{k=10}$  &  DilBert$_{k=11}$  &  AlBert  &  DilAlBert$_{k=10}$  &  DilAlBert$_{k=11}$\\ 
\midrule
  &  NI Q  &    &1.0 (1e-3)&0.9 (8.2e-4)&    &1.2 (1.2e-3)&  1.4 (1.3e-3)\\
GPU  &  NI P  &   &0.8 (8e-4)&1.1 (1e-3)&    &1.0 (1.0e-3)&  1.1 (1.0e-3)\\ 
  &  I Q-P  &117.9 (9.8e-4)&17.0 (8.5e-4)&9.9 (9.9e-4)&142.2 (1.2e-3)&22.4 (1.1e-3)& 13.1 (1.3e-3)\\ 
  & Total &117.9&18.8 &11.9&142.2&24.6& 15.6 \\
  &  Speedup  &  x1  &x6.3&x9.9& x1 &x5.8&  x9.1\\ 
  \midrule
  &  NI Q  &&  3.5 (3.5e-3)  & 3.9 (3.6e-3)  &    &4.1 (4.1e-3)&  4.8 (4.3e-3)\\ 
10-threads  &  NI P  &    & 12.0 (1.2e-2)&  13.34 (1.2e-2)  &    &16.7 (1.7e-2)&    17.6 (1.6e-2)\\ 
   CPU &  I Q-P  &2602.6 (2.2e-2)&258.7 (1.3e-2)  &  126.7 (1.3e-2)  & 2597.6  (2.2e-2)&346.8 (1.7e-2)&  176.3 (1.7e-2)\\ 
  &Total&2602.6&274.2&144.0&2597.6&367.7&198.7 \\
&  Speedup  &  x1  &  x9.5&  x18.1  &  x1  &  x7,1 &  x13.1 \\
  \bottomrule
\end{tabular}
}\caption{Time consumption to search for the answer of 100 questions in 100 passages with Bert, Albert, DilBert and DilAlbert on GPU and on a 10-threads CPU. "NI Q" (resp. "NI P") refers to the time required to process the $q$ questions (resp the $p$ passages) in the Non Interaction layers, and "I Q-P" refers to time required to process the $p\times q$ pairs in the Interaction layers. We also report in brackets a time consumption normalized with respect to the number of blocs ($l$, $k$ or $l-k$) and inputs ($p$, $q$ or $p\times q$).}
\label{time_consumption_encoder_dilencoder}
\end{table*}

Experiments are carried on a bull server with a Nvidia Tesla V100 GPU and a 5-cores/10-threads Intel Xeon Gold 6132 (2.6-3.7 GHz) CPU. 

The empirical results confirm the theoretical intuitions of the \textit{Complexity} paragraph. Firstly, since in-bloc computations are parallelizable and a GPU device has a great parallelization power, we can observe on GPU that the processing time per bloc per input sequence is approximately always the same (around 1e-3) whereas it depends more on the sequence length on CPU (around 4e-3 for the question with a length of $n_q \approx 16$ and 1.3e-2 for the question-paragraph concatenation with a length of $n_s = 384$). Secondly, in such an ODQA setting where the answers of multiple ($q$) questions will be searched in multiple paragraphs ($p$), computations that depend on couples of question-passage (i.e. in interaction blocs for Dil variants and in all blocs for original models) take most of the time because they are proportional to the number of pairs $p\times q$ which is largely superior to $p$ and $q$. Since original models have $l=12$ interaction layers and Dil variants only have $l-k$ ($1$ or $2$) interaction layers, it results that the speedup factor in the encoder entailed by delayed interaction tend to $\frac{l}{l-k}$ for a large number of questions and passages. We therefore empirically observe a speedup of around one order of magnitude for Dil variants in Table \ref{time_consumption_encoder_dilencoder}.
\subsection{Open Domain QA}

Apart from speedup, it is necessary to check whether delayed interaction has an impact on the quality of answers in Open Domain QA. Therefore, we finally focus here, as \citet{chen2017reading}, on the ODQA challenge referred to as OpenSQuAD. It consists in confronting algorithms to the task of answering to questions using the entire English Wikipedia\footnote{We consider a dump from December 2016 as proposed by \citet{chen2017reading}.}. More precisely, it considers the 10570 questions from the SQuAD v1.1 dev set but does not provide the associated paragraph containing the answer so the algorithms have to search in a larger set of 5,075,182 articles.

\begin{figure}[t]
\centering
\includegraphics[width=0.95\columnwidth]{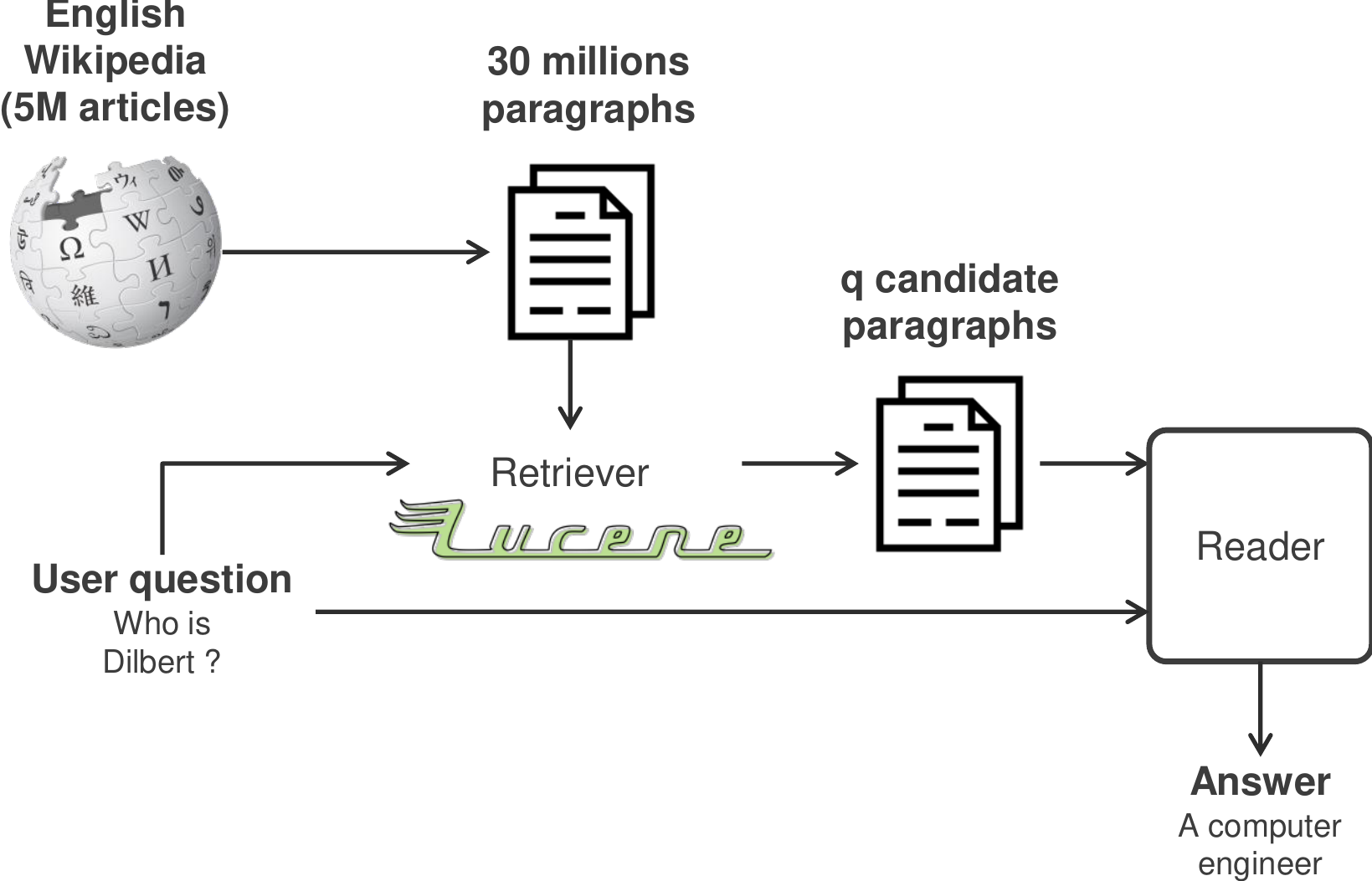}
\caption{General architecture of the question answering pipeline used in our experiments.}
\label{odqa_pipeline}
\end{figure}

Since our proposal specifically focuses on the reader part of the classical retriever + reader ODQA pipeline, we mainly analyze its impact on the global performance. More precisely, for the remaining parts, we will arbitrarily choose a proven baseline approach from the literature as long as it relies on a transformer-based reader, and we will simply substitute the latter with the model that we want to evaluate. The baseline BertSerini \cite{yang2019end} is a good candidate framework for our experiment as it uses a robust retriever and Bert as a reader. Since its implementation is not available, we detail ours below (Figure \ref{odqa_pipeline}).  

\paragraph{ODQA Pipeline}

We first apply a preprocessing step to properly index the Wikipedia dump with Lucene using the pyserini library, version 0.9.4.0 \cite{yang2017anserini,yang2018anserini}. As Wikipedia articles are rather long and sometimes cover multiple topics, they are difficult to process as such for both the retriever and the reader \cite{yang2019end,wang2017r,lee2018ranking}, so we start by splitting them into paragraphs. We consider two strategies: 
\begin{itemize}
    \item Using a double newline as a delimiter. We then discard all items with less than 30 characters leading to a number of paragraphs of around 29.1 millions approximately as in \cite{yang2019end}.
    \item Using a fixed length of 100 words with a sliding window of 50 words as in \cite{wang2019multi}.
\end{itemize}

We obtain close end-to-end results with both techniques but retain the slightly better second strategy for the final pipeline. The splitted paragraphs are then processed with the pyserini script \textit{index}\footnote{\url{https://github.com/castorini/pyserini}} which outputs all the necessary inverted indices for retrieval.

Once preprocessing is over, we can start making predictions. The retriever is a pyserini SimpleSearcher that relies on the computed inverted indices, and that uses BM25 (pyserini's default parameters) as a relevance score. For each question, it returns the $p$ paragraphs with the top score $s_{bm25}$. We choose $p=29$ and $p=100$ as in BertSerini. This retrieval step takes less than 0.15s on CPU with the hardware configuration described in the previous section. Then, the reader is applied to each paragraph and produces a start logit score $s_s$ and an end logit score $s_e$ for each token. To aggregate the results into a final prediction, we consider two variants: one that excludes the retriever score and selects the text span with the highest reader score $s_r = \frac{s_s + s_e}{2}$ across all paragraphs, and one that includes it by replacing $s_r$ with $\mu s_r + (1-\mu)s_{bm25}$ where $\mu$ is a hyperparameter between 0 and 1. A cross validation on a subset of the SQuAD v1.1 train set questions shows that, from 0.1 to 0.9 in steps of 0.1, the value of $\mu = 0.5$ leads to an optimal end-to-end performance for all considered readers. The latter are the trained models (Bert, DilBert$_{k=10}$, Albert and DilAlbert$_{k=10}$) from section \textit{eQA performance}. We add the multilingual version of Bert \cite{pires2019multilingual,devlin2019bert} as it is currently by far the most downloaded model from the Huggingface platform.

\paragraph{Results} 

We evaluate the ODQA pipeline with the exact match (EM) and the F1-score (F1) between the text predicted from the Wikipedia dump and the expected answer. Obviously, the task here is more difficult than eQA since the paragraph containing the answer is not provided, so the performance is excepted to be lower than in Figure \ref{evo_performance_dilbert} and \ref{evo_performance_dilalbert}. Not only the IR part will select $p$ paragraphs that might not contain the answer, but additional difficulties that we discuss in the next section emerge when the search space is large and diverse. To give an idea of the impact of the retriever part on the end-to-end result, we compute the percentage of questions for which the expected answer at least appears in the $p$ selected paragraphs (column $R$ in Table \ref{table:main}). Since the paragraph splitting methodology and the version of pyserini were not detailed by \citet{yang2019end} we were only able to obtain the same IR performance as Bertserini for $p = 29$ but we obtained a lower $R$ for $p=100$.

\begin{table}[t]
\centering\resizebox{\columnwidth}{!}{
\begin{tabular}{lccccc}
\toprule
Model 				& \multicolumn{2}{c}{EM} &  \multicolumn{2}{c}{F1}   & R  \\ 
& w/o & w/ & w/o & w/ & \\
\midrule
DrQA \cite{chen2017reading}	        & 27.1 & - & - &  -  & 77.8 \\
$R^3$ \cite{wang2017r}   	& - &29.1 & - & 37.5 & -    \\
Par. R. \cite{lee2018ranking}         & - & 28.5 & - & -     & 83.1 \\
\textsc{Minimal}~\cite{min2018efficient}         &  - & 34.7 & - & 42.5 & 64.0 \\
\midrule
\citet{yang2019end}: & & & \\
Bertserini ($p=29$)  & - & 36.6 & - & 44.0 & 75.0 \\
Bertserini ($p=100$) & - & 38.6 & - & 46.1 & 85.8 \\
\midrule
Ours: & & & \\
Bert  ($p=29$)  & 37.4 & 43.5 & 44.1 & 50.9 & 74.6 \\
DilBert  ($p=29$)  & 38.8 & 42.0 & 46.7 & 50.4 & 74.6 \\
\cmidrule{2-6}
Bert  ($p=100$)  & 31.8 & 44.3 & 38.0 & 51.8 & 82.4 \\
DilBert  ($p=100$)  & 36.5 & 43.2 & 43.8 & 51.6 & 82.4 \\
\cmidrule{2-6}
mBert  ($p=29$)  & 34.4 & 41.9 & 40.6 & 49.1  & 74.6 \\
DilmBert  ($p=29$)  & 38.5 & 42.8 & 45.8 & 50.5 & 74.6 \\
\cmidrule{2-6}
mBert  ($p=100$)  & 28.3 & 42.3 & 34.1 & 49.3  & 82.4 \\
DilmBert  ($p=100$)  & 34.7 & 43.8 & 41.7 & 51.6 & 82.4 \\
\cmidrule{2-6}
Albert  ($p=29$) & 40.6 & 44.1 & 47.0 & 51.3 &   74.6  \\
DilAlbert  ($p=29$)  & 39.7 & 43.6 & 47.7 & 51.8 & 74.6 \\
\cmidrule{2-6}
Albert  ($p=100$)  & 36.2 & 44.9 & 42.1 & 52.1 & 82.4 \\
DilAlbert  ($p=100$)  & 36.8 & 44.8 & 43.8 & 52.9 & 82.4 \\
\bottomrule
\end{tabular}}
\caption{End-to-end performance (EM, F1) of our ODQA pipeline with several models as readers (Bert, DilBert, multilingual Bert or mBert, DilmBert, Albert and DilAlbert). Results when excluding the IR score (w/o) and when including it (w/) are detailed. Previous results from the literature are also reported above.}
\label{table:main}
\end{table}

The final end-to-end ODQA results exhibit interesting properties (columns EM/F1 in Table  \ref{table:main}). First we can notice that, although the variants with delayed interaction were a little worse than original models on the eQA task (e.g. Figure \ref{evo_performance_dilbert}), they almost always outperform them for ODQA when the IR score is not used, and sometimes by a significant margin (e.g. when there is a high number of candidate paragraphs $p$). When exploiting the IR score, the conclusion are more mixed. A potential explanation is that delayed interaction acts like a sort of regularization similarly to dropout \cite{srivastava2014dropout}, in the sense that the resulting architecture is the same as the original one except that some connections are removed in the first $k$ blocs. Therefore, the proposed variants are more prone to generalization and produce less noisy predictions from passages with a low IR relevance score, whereas original models tend to do so \cite{yang2019end,xie2020distant}. Using the IR score, all models are able to focus on their predictions from the most relevant passages which helps improving the overall performance. In this case, original models manage to catch up with delayed interaction variants. 

The results with the multilingual weights are slightly worse than with the English ones, but mostly for the Bert model. In fact, the variant DilmBert is always better than the original mBert. As for the results with Albert, they are close to those of Bert but a little better in general.

\begin{table*}[t]
\centering\resizebox{\textwidth}{!}{
\begin{tabular}{ccc}
\toprule
Question & Expected Answer & Predicted Answer \\
\midrule
When did Zwilling and Karistadt become active at Wittenberg? & June 1521 & mid-1521 \\
The Los Angeles Angels of Anaheim are from which sport? & MLB & Major League Baseball \\
How many fumbles did Von Miller force in Super Bowl 50? & 2 & two \\
What is the AFC short for? & American Football Conference & Asian Football Confederation \\
How many graduate students does Harvard have? & 14000 & 15000 \\
What position did Newton play during Super Bowl 50? & quarterback & QB \\
\bottomrule
\end{tabular}}
\caption{Examples of questions, expected answers, and answers predicted by an ODQA pipeline on OpenSQuAD.}
\label{table:odqa_example}
\end{table*}

On a final side note, the best performance (EM: 44.3, F1: 51.8) obtained in this study with a Bert Base English uncased, simply fine-tuned for the eQA task on SQuAD v1.1 and combined with BM25, is significantly better than that of Bertserini while the latter uses the same building blocks, and have a higher IR score (R). This improvement is mainly due to engineering (documents splitting, Lucene's version), and it does not surprise us as a similar observation was made in the recent \cite{xie2020distant} which reports updated higher results for BertSerini (EM: 41.8, F1: 49.5, R: 86.3). 

Our update (EM:44.3, F1:51.8) continues to bring an accessible baseline BM25 + Bert Base closer to the state-of-the-art results on OpenSQuAD which, to the best of our knowledge, are achieved by Multi-Passage Bert (EM:51.2, F1:59.0). The latter exploits additional mechanisms to improve the results: a Bert-based reranker for the IR part, and a global normalization process, for the predictions of the reader, specifically trained in a multi-passage setting. 

\section{Discussion}

Our study shows that interesting opportunities arise when tackling Open Domain Question Answering from the angle of the reader's architecture. Our initial objective was to exploit specificities in the nature of the ODQA problem to reduce computations. The complexity of algorithms is a major industrial and societal issue in today's research as it controls the efficiency of systems in production but also their carbon footprint \cite{thakur2016towards}. As shown in Table \ref{time_consumption_encoder_dilencoder}, the proposed variants, with $k=10$ allow to reduce the number of interaction blocs from $12$ to $2$ in the widely spread Bert and Albert, which entails a 85\% (6x) speedup in the ODQA setting. The approach is generic: it can be applied to several transformer-based architectures, and still benefit from their available sets of model weights (e.g. English uncased, multilingual) to avoid the costly pre-training. It is also compatible with standard retriever + reader ODQA pipeline in several ways. We considered in this work a framework where a transformer-based model is used as the reader part because it is the case for the entirety of recent ODQA proposals. In some of them, the retriever/reranker is also based on a language model \cite{wang2019multi}, so delayed interaction could be applied to this part of the pipeline as well. 

In addition to having achieved our initial speedup objective, the proposed variants have also proven to be highly competitive in terms of quality of answer for Open Domain Question Answering. In particular, they can outperform original models when it comes to search for the answer in a large variety of passages. This could be a matter of generalization. The SQuAD v1.1 dataset is based on a few hundreds of Wikipedia articles, hence the measured eQA performance on the dev set can be partly due to overspecialization on a few topics \cite{xie2020distant}. The regularization entailed by delayed interaction reduces the performance for the most relevant passage (Figure \ref{evo_performance_dilbert} and \ref{evo_performance_dilalbert}) but makes up for it by reducing unexpected behaviour on irrelevant passages (Table \ref{table:main}). Solutions to help on that issue with Bert have been recently explored \cite{wang2019multi,xie2020distant} and they consist in using global normalization or distant supervision with irrelevant passages during training to make the model more robust to a multiple passage setting. These interesting ideas allow a significant improvement of the performance and they do not exclude the possibility of using DilBert instead of Bert, at least to benefit from its speedup.

One last important point to mention is the current absolute state-of-the-art results on the ODQA dataset OpenSQuAD. Although they appear low (especially in comparison with the eQA setting), they can be explained by challenges and biases in the OpenSQuAD dataset and the associated metrics. Apart from the IR bottleneck that entails a 15\%-18\% unavoidable loss for the reader, there are questions from the SQuAD v1.1 dev set that (see Table \ref{table:odqa_example}):
\begin{enumerate}
    \item simply become unanswerable in OpenSQuAD because of their ambiguity. For instance, the question "What is the AFC short for?" needs to be associated to a specific paragraph to have a specific meaning. 
    \item have several valid answers, located in multiple paragraphs and under different forms. It mostly affects the exact match but sometimes the F1-score is also penalized: for instance, for the question "What position did Newton play during Super Bowl 50?", although the model finds the correct answer "QB", its recall is really low because it is expected to predict "quarterback".
\end{enumerate}

\section{Conclusion}

In this work we used a Delayed Interaction mechanism in transformer-based RC models which allowed a performance improvement and a speedup of up to an order of magnitude. It is rather generic and can benefit from existing resources (architectures, pre-trained weights). \textit{A posteriori}, we could establish a connection with very recent proposals such as Big Bird \cite{zaheer2020big} and more generally, with a global strategy of restricted/partial attention which tends to develop in current research trends. Our study allows to show the effectiveness of partial attention for ODQA but there remains many paths to explore. One of them is to improve the training by (i) introducing non interaction layers in the pre-training task and (ii) following the ideas of \citet{xie2020distant,wang2019multi} for a multi-passage fine-tuning. Another interesting direction is to address the memory management issue in the ODQA problem. In particular, storing pre-computed representations of documents for either IR or eQA can become costly for large scale databases, and it could be interesting to explore the impact of compressing them with sparsification \cite{sun2016sparse} or binarization \cite{tissier2019near} for instance.
 
\bibliography{biblio.bib}

\begin{thebibliography}{48}
\providecommand{\natexlab}[1]{#1}
\providecommand{\url}[1]{\texttt{#1}}
\providecommand{\urlprefix}{URL }
\expandafter\ifx\csname urlstyle\endcsname\relax
  \providecommand{\doi}[1]{doi:\discretionary{}{}{}#1}\else
  \providecommand{\doi}{doi:\discretionary{}{}{}\begingroup
  \urlstyle{rm}\Url}\fi

\bibitem[{Baeza-Yates, Ribeiro-Neto et~al.(1999)}]{baeza1999modern}
Baeza-Yates, R.; Ribeiro-Neto, B.; et~al. 1999.
\newblock \emph{Modern information retrieval}, volume 463.
\newblock ACM press New York.

\bibitem[{Bia{\l}ecki et~al.(2012)Bia{\l}ecki, Muir, Ingersoll, and
  Imagination}]{bialecki2012apache}
Bia{\l}ecki, A.; Muir, R.; Ingersoll, G.; and Imagination, L. 2012.
\newblock Apache lucene 4.
\newblock In \emph{SIGIR 2012 workshop on open source information retrieval},
  17.

\bibitem[{B{\"u}ttcher, Clarke, and Cormack(2016)}]{buttcher2016information}
B{\"u}ttcher, S.; Clarke, C.~L.; and Cormack, G.~V. 2016.
\newblock \emph{Information retrieval: Implementing and evaluating search
  engines}.
\newblock Mit Press.

\bibitem[{Chen, Bolton, and Manning(2016)}]{chen2016thorough}
Chen, D.; Bolton, J.; and Manning, C.~D. 2016.
\newblock A Thorough Examination of the CNN/Daily Mail Reading Comprehension
  Task.
\newblock In \emph{Proceedings of the 54th Annual Meeting of the Association
  for Computational Linguistics (Volume 1: Long Papers)}, 2358--2367.

\bibitem[{Chen et~al.(2017)Chen, Fisch, Weston, and Bordes}]{chen2017reading}
Chen, D.; Fisch, A.; Weston, J.; and Bordes, A. 2017.
\newblock Reading Wikipedia to Answer Open-Domain Questions.
\newblock In \emph{Proceedings of the 55th Annual Meeting of the Association
  for Computational Linguistics (Volume 1: Long Papers)}, 1870--1879.

\bibitem[{Clark et~al.(2019{\natexlab{a}})Clark, Khandelwal, Levy, and
  Manning}]{clark2019does}
Clark, K.; Khandelwal, U.; Levy, O.; and Manning, C.~D. 2019{\natexlab{a}}.
\newblock What Does BERT Look at? An Analysis of BERT’s Attention.
\newblock In \emph{Proceedings of the 2019 ACL Workshop BlackboxNLP: Analyzing
  and Interpreting Neural Networks for NLP}, 276--286.

\bibitem[{Clark et~al.(2019{\natexlab{b}})Clark, Luong, Le, and
  Manning}]{clark2019electra}
Clark, K.; Luong, M.-T.; Le, Q.~V.; and Manning, C.~D. 2019{\natexlab{b}}.
\newblock ELECTRA: Pre-training Text Encoders as Discriminators Rather Than
  Generators.
\newblock In \emph{International Conference on Learning Representations}.

\bibitem[{Dai et~al.(2018)Dai, Xiong, Callan, and Liu}]{dai2018convolutional}
Dai, Z.; Xiong, C.; Callan, J.; and Liu, Z. 2018.
\newblock Convolutional neural networks for soft-matching n-grams in ad-hoc
  search.
\newblock In \emph{Proceedings of the eleventh ACM international conference on
  web search and data mining}, 126--134.

\bibitem[{Das et~al.(2016)Das, Yenala, Chinnakotla, and
  Shrivastava}]{das2016together}
Das, A.; Yenala, H.; Chinnakotla, M.; and Shrivastava, M. 2016.
\newblock Together we stand: Siamese networks for similar question retrieval.
\newblock In \emph{Proceedings of the 54th Annual Meeting of the Association
  for Computational Linguistics (Volume 1: Long Papers)}, 378--387.

\bibitem[{Devlin et~al.(2019)Devlin, Chang, Lee, and
  Toutanova}]{devlin2019bert}
Devlin, J.; Chang, M.-W.; Lee, K.; and Toutanova, K. 2019.
\newblock BERT: Pre-training of Deep Bidirectional Transformers for Language
  Understanding.
\newblock In \emph{Proceedings of the 2019 Conference of the North American
  Chapter of the Association for Computational Linguistics: Human Language
  Technologies, Volume 1 (Long and Short Papers)}, 4171--4186.

\bibitem[{Feldman and El-Yaniv(2019)}]{feldman2019multi}
Feldman, Y.; and El-Yaniv, R. 2019.
\newblock Multi-hop paragraph retrieval for open-domain question answering.
\newblock \emph{arXiv preprint arXiv:1906.06606} .

\bibitem[{Ganguly et~al.(2015)Ganguly, Roy, Mitra, and Jones}]{ganguly2015word}
Ganguly, D.; Roy, D.; Mitra, M.; and Jones, G.~J. 2015.
\newblock Word embedding based generalized language model for information
  retrieval.
\newblock In \emph{Proceedings of the 38th international ACM SIGIR conference
  on research and development in information retrieval}, 795--798.

\bibitem[{Guo et~al.(2016)Guo, Fan, Ai, and Croft}]{guo2016deep}
Guo, J.; Fan, Y.; Ai, Q.; and Croft, W.~B. 2016.
\newblock A deep relevance matching model for ad-hoc retrieval.
\newblock In \emph{Proceedings of the 25th ACM International on Conference on
  Information and Knowledge Management}, 55--64.

\bibitem[{Hiemstra(2000)}]{hiemstra2000probabilistic}
Hiemstra, D. 2000.
\newblock A probabilistic justification for using tf$\times$ idf term weighting
  in information retrieval.
\newblock \emph{International Journal on Digital Libraries} 3(2): 131--139.

\bibitem[{Lan et~al.(2019)Lan, Chen, Goodman, Gimpel, Sharma, and
  Soricut}]{lan2019albert}
Lan, Z.; Chen, M.; Goodman, S.; Gimpel, K.; Sharma, P.; and Soricut, R. 2019.
\newblock Albert: A lite bert for self-supervised learning of language
  representations.
\newblock \emph{arXiv preprint arXiv:1909.11942} .

\bibitem[{Lee et~al.(2018)Lee, Yun, Kim, Ko, and Kang}]{lee2018ranking}
Lee, J.; Yun, S.; Kim, H.; Ko, M.; and Kang, J. 2018.
\newblock Ranking paragraphs for improving answer recall in open-domain
  question answering.
\newblock \emph{arXiv preprint arXiv:1810.00494} .

\bibitem[{Lee, Chang, and Toutanova(2019)}]{lee2019latent}
Lee, K.; Chang, M.-W.; and Toutanova, K. 2019.
\newblock Latent retrieval for weakly supervised open domain question
  answering.
\newblock \emph{arXiv preprint arXiv:1906.00300} .

\bibitem[{Lin(2019)}]{lin2019neural}
Lin, J. 2019.
\newblock The neural hype and comparisons against weak baselines.
\newblock In \emph{ACM SIGIR Forum}, volume~52, 40--51. ACM New York, NY, USA.

\bibitem[{Liu et~al.(2019)Liu, Ott, Goyal, Du, Joshi, Chen, Levy, Lewis,
  Zettlemoyer, and Stoyanov}]{liu2019roberta}
Liu, Y.; Ott, M.; Goyal, N.; Du, J.; Joshi, M.; Chen, D.; Levy, O.; Lewis, M.;
  Zettlemoyer, L.; and Stoyanov, V. 2019.
\newblock Roberta: A robustly optimized bert pretraining approach.
\newblock \emph{arXiv preprint arXiv:1907.11692} .

\bibitem[{MacAvaney et~al.(2019)MacAvaney, Yates, Cohan, and
  Goharian}]{macavaney2019cedr}
MacAvaney, S.; Yates, A.; Cohan, A.; and Goharian, N. 2019.
\newblock CEDR: Contextualized embeddings for document ranking.
\newblock In \emph{Proceedings of the 42nd International ACM SIGIR Conference
  on Research and Development in Information Retrieval}, 1101--1104.

\bibitem[{Manning, Sch{\"u}tze, and Raghavan(2008)}]{manning2008introduction}
Manning, C.~D.; Sch{\"u}tze, H.; and Raghavan, P. 2008.
\newblock \emph{Introduction to information retrieval}.
\newblock Cambridge university press.

\bibitem[{Mikolov et~al.(2013)Mikolov, Sutskever, Chen, Corrado, and
  Dean}]{mikolov2013distributed}
Mikolov, T.; Sutskever, I.; Chen, K.; Corrado, G.~S.; and Dean, J. 2013.
\newblock Distributed representations of words and phrases and their
  compositionality.
\newblock In \emph{Advances in neural information processing systems},
  3111--3119.

\bibitem[{Min et~al.(2018)Min, Zhong, Socher, and Xiong}]{min2018efficient}
Min, S.; Zhong, V.; Socher, R.; and Xiong, C. 2018.
\newblock Efficient and Robust Question Answering from Minimal Context over
  Documents.
\newblock In \emph{Proceedings of the 56th Annual Meeting of the Association
  for Computational Linguistics (Volume 1: Long Papers)}, 1725--1735.

\bibitem[{Mitra, Craswell et~al.(2018)}]{mitra2018introduction}
Mitra, B.; Craswell, N.; et~al. 2018.
\newblock \emph{An introduction to neural information retrieval}.
\newblock Now Foundations and Trends.

\bibitem[{Nogueira and Cho(2019)}]{nogueira2019passage}
Nogueira, R.; and Cho, K. 2019.
\newblock Passage Re-ranking with BERT.
\newblock \emph{arXiv preprint arXiv:1901.04085} .

\bibitem[{Pires, Schlinger, and Garrette(2019)}]{pires2019multilingual}
Pires, T.; Schlinger, E.; and Garrette, D. 2019.
\newblock How Multilingual is Multilingual BERT?
\newblock In \emph{Proceedings of the 57th Annual Meeting of the Association
  for Computational Linguistics}, 4996--5001.

\bibitem[{Raffel et~al.(2019)Raffel, Shazeer, Roberts, Lee, Narang, Matena,
  Zhou, Li, and Liu}]{raffel2019exploring}
Raffel, C.; Shazeer, N.; Roberts, A.; Lee, K.; Narang, S.; Matena, M.; Zhou,
  Y.; Li, W.; and Liu, P.~J. 2019.
\newblock Exploring the limits of transfer learning with a unified text-to-text
  transformer.
\newblock \emph{arXiv preprint arXiv:1910.10683} .

\bibitem[{Rajpurkar et~al.(2016)Rajpurkar, Zhang, Lopyrev, and
  Liang}]{rajpurkar2016squad}
Rajpurkar, P.; Zhang, J.; Lopyrev, K.; and Liang, P. 2016.
\newblock SQuAD: 100,000+ Questions for Machine Comprehension of Text.
\newblock In \emph{Proceedings of the 2016 Conference on Empirical Methods in
  Natural Language Processing}, 2383--2392.

\bibitem[{Ren, Cheng, and Su(2020)}]{ren2020multi}
Ren, Q.; Cheng, X.; and Su, S. 2020.
\newblock Multi-Task Learning with Generative Adversarial Training for
  Multi-Passage Machine Reading Comprehension.
\newblock In \emph{AAAI}, 8705--8712.

\bibitem[{Robertson et~al.(1995)Robertson, Walker, Jones, Hancock-Beaulieu,
  Gatford et~al.}]{robertson1995okapi}
Robertson, S.~E.; Walker, S.; Jones, S.; Hancock-Beaulieu, M.~M.; Gatford, M.;
  et~al. 1995.
\newblock Okapi at TREC-3.
\newblock \emph{Nist Special Publication Sp} 109: 109.

\bibitem[{Ryu, Jang, and Kim(2014)}]{ryu2014open}
Ryu, P.-M.; Jang, M.-G.; and Kim, H.-K. 2014.
\newblock Open domain question answering using Wikipedia-based knowledge model.
\newblock \emph{Information Processing \& Management} 50(5): 683--692.

\bibitem[{Srivastava et~al.(2014)Srivastava, Hinton, Krizhevsky, Sutskever, and
  Salakhutdinov}]{srivastava2014dropout}
Srivastava, N.; Hinton, G.; Krizhevsky, A.; Sutskever, I.; and Salakhutdinov,
  R. 2014.
\newblock Dropout: a simple way to prevent neural networks from overfitting.
\newblock \emph{The journal of machine learning research} 15(1): 1929--1958.

\bibitem[{Sun et~al.(2016)Sun, Guo, Lan, Xu, and Cheng}]{sun2016sparse}
Sun, F.; Guo, J.; Lan, Y.; Xu, J.; and Cheng, X. 2016.
\newblock Sparse word embeddings using l1 regularized online learning.
\newblock In \emph{Proceedings of the Twenty-Fifth International Joint
  Conference on Artificial Intelligence}, 2915--2921. AAAI Press.

\bibitem[{Thakur and Chaurasia(2016)}]{thakur2016towards}
Thakur, S.; and Chaurasia, A. 2016.
\newblock Towards Green Cloud Computing: Impact of carbon footprint on
  environment.
\newblock In \emph{2016 6th International Conference-Cloud System and Big Data
  Engineering (Confluence)}, 209--213. IEEE.

\bibitem[{Tissier, Gravier, and Habrard(2019)}]{tissier2019near}
Tissier, J.; Gravier, C.; and Habrard, A. 2019.
\newblock Near-lossless binarization of word embeddings.
\newblock In \emph{Proceedings of the AAAI Conference on Artificial
  Intelligence}, volume~33, 7104--7111.

\bibitem[{Vaswani et~al.(2017)Vaswani, Shazeer, Parmar, Uszkoreit, Jones,
  Gomez, Kaiser, and Polosukhin}]{vaswani2017attention}
Vaswani, A.; Shazeer, N.; Parmar, N.; Uszkoreit, J.; Jones, L.; Gomez, A.~N.;
  Kaiser, L.; and Polosukhin, I. 2017.
\newblock Attention Is All You Need .

\bibitem[{Wang et~al.(2018)Wang, Singh, Michael, Hill, Levy, and
  Bowman}]{wang2018glue}
Wang, A.; Singh, A.; Michael, J.; Hill, F.; Levy, O.; and Bowman, S.~R. 2018.
\newblock Glue: A multi-task benchmark and analysis platform for natural
  language understanding.
\newblock \emph{arXiv preprint arXiv:1804.07461} .

\bibitem[{Wang et~al.(2017)Wang, Yu, Guo, Wang, Klinger, Zhang, Chang, Tesauro,
  Zhou, and Jiang}]{wang2017r}
Wang, S.; Yu, M.; Guo, X.; Wang, Z.; Klinger, T.; Zhang, W.; Chang, S.;
  Tesauro, G.; Zhou, B.; and Jiang, J. 2017.
\newblock $R^3$: Reinforced reader-ranker for open-domain question answering.
\newblock \emph{arXiv preprint arXiv:1709.00023} .

\bibitem[{Wang et~al.(2019)Wang, Ng, Ma, Nallapati, and Xiang}]{wang2019multi}
Wang, Z.; Ng, P.; Ma, X.; Nallapati, R.; and Xiang, B. 2019.
\newblock Multi-passage bert: A globally normalized bert model for open-domain
  question answering.
\newblock \emph{arXiv preprint arXiv:1908.08167} .

\bibitem[{Wolf et~al.(2019)Wolf, Debut, Sanh, Chaumond, Delangue, Moi, Cistac,
  Rault, Louf, Funtowicz, Davison, Shleifer, von Platen, Ma, Jernite, Plu, Xu,
  Scao, Gugger, Drame, Lhoest, and Rush}]{wolf2019huggingfaces}
Wolf, T.; Debut, L.; Sanh, V.; Chaumond, J.; Delangue, C.; Moi, A.; Cistac, P.;
  Rault, T.; Louf, R.; Funtowicz, M.; Davison, J.; Shleifer, S.; von Platen,
  P.; Ma, C.; Jernite, Y.; Plu, J.; Xu, C.; Scao, T.~L.; Gugger, S.; Drame, M.;
  Lhoest, Q.; and Rush, A.~M. 2019.
\newblock HuggingFace's Transformers: State-of-the-art Natural Language
  Processing.

\bibitem[{Woods and WA(1977)}]{woods1977lunar}
Woods, W.~A.; and WA, W. 1977.
\newblock Lunar rocks in natural English: Explorations in natural language
  question answering. .

\bibitem[{Xie et~al.(2020)Xie, Yang, Tan, Xiong, Yuan, Huai, Li, and
  Lin}]{xie2020distant}
Xie, Y.; Yang, W.; Tan, L.; Xiong, K.; Yuan, N.~J.; Huai, B.; Li, M.; and Lin,
  J. 2020.
\newblock Distant Supervision for Multi-Stage Fine-Tuning in Retrieval-Based
  Question Answering.
\newblock In \emph{Proceedings of The Web Conference 2020}, 2934--2940.

\bibitem[{Yang, Fang, and Lin(2017)}]{yang2017anserini}
Yang, P.; Fang, H.; and Lin, J. 2017.
\newblock Anserini: Enabling the use of Lucene for information retrieval
  research.
\newblock In \emph{Proceedings of the 40th International ACM SIGIR Conference
  on Research and Development in Information Retrieval}, 1253--1256.

\bibitem[{Yang, Fang, and Lin(2018)}]{yang2018anserini}
Yang, P.; Fang, H.; and Lin, J. 2018.
\newblock Anserini: Reproducible ranking baselines using Lucene.
\newblock \emph{Journal of Data and Information Quality (JDIQ)} 10(4): 1--20.

\bibitem[{Yang et~al.(2019{\natexlab{a}})Yang, Xie, Lin, Li, Tan, Xiong, Li,
  and Lin}]{yang2019end}
Yang, W.; Xie, Y.; Lin, A.; Li, X.; Tan, L.; Xiong, K.; Li, M.; and Lin, J.
  2019{\natexlab{a}}.
\newblock End-to-end open-domain question answering with bertserini.
\newblock \emph{arXiv preprint arXiv:1902.01718} .

\bibitem[{Yang, Zhang, and Lin(2019)}]{yang2019simple}
Yang, W.; Zhang, H.; and Lin, J. 2019.
\newblock Simple applications of BERT for ad hoc document retrieval.
\newblock \emph{arXiv preprint arXiv:1903.10972} .

\bibitem[{Yang et~al.(2019{\natexlab{b}})Yang, Dai, Yang, Carbonell,
  Salakhutdinov, and Le}]{yang2019xlnet}
Yang, Z.; Dai, Z.; Yang, Y.; Carbonell, J.; Salakhutdinov, R.~R.; and Le, Q.~V.
  2019{\natexlab{b}}.
\newblock Xlnet: Generalized autoregressive pretraining for language
  understanding.
\newblock In \emph{Advances in neural information processing systems},
  5753--5763.

\bibitem[{Zaheer et~al.(2020)Zaheer, Guruganesh, Dubey, Ainslie, Alberti,
  Ontanon, Pham, Ravula, Wang, Yang et~al.}]{zaheer2020big}
Zaheer, M.; Guruganesh, G.; Dubey, A.; Ainslie, J.; Alberti, C.; Ontanon, S.;
  Pham, P.; Ravula, A.; Wang, Q.; Yang, L.; et~al. 2020.
\newblock Big bird: Transformers for longer sequences.
\newblock \emph{arXiv preprint arXiv:2007.14062} .

\end{thebibliography}

\end{document}